%% file: arxiv.tex
\documentclass[mlmain]{jmlr}

\usepackage{longtable}

\usepackage{booktabs}
\usepackage[load-configurations=version-1]{siunitx} 

\theorembodyfont{\upshape}
\theoremheaderfont{\scshape}
\theorempostheader{:}
\theoremsep{\newline}

\jmlrvolume{}
\firstpageno{1}

\jmlryear{2026}
\jmlrworkshop{Symmetry and Geometry in Neural Representations}

\usepackage[most]{tcolorbox}
\usepackage{xcolor}
\newcommand{\cmark}{\textcolor{green!60!black}{\checkmark}}

\title[Data-driven Lie symmetries with latent ODEs]{Data-driven Discovery of Lie Symmetries and First Integrals with Latent ODEs}

\author[Wiemann \& Melchior]{\Name{Matt L. Wiemann}\\ \Email{matt.sampson@princeton.edu} \\ \addr{Princeton University} \\ 
\Name{Peter Melchior}\\ \addr{Princeton University}}

\begin{document}

\maketitle
\thispagestyle{empty}

\begin{abstract}
We present \textsc{LatentLieODE}, a novel neural ODE method to discover linear Lie symmetries and first integrals of the learned dynamics from timeseries observations such as video inputs. It starts with an autoencoder, which learns a low dimensional representation of the observed system and then maps trajectories to decorrelated latent coordinates $z\in\mathbb{R}^d$. A bank of linear infinitesimal generators $\mathbf{A}_k \in \mathbb{R}^{d\times d}$ is fitted jointly against a differentiable Lie-bracket residual $\mathbf{J}_f(z)\mathbf{A}z - \mathbf{A} f(z)$, while an MLP optimises the latent ODE $f(z)=\mathrm{d}{z}/\mathrm{d}t$ to reconstruct the observation-space timeseries. After suitable generators and $f$ are found, we fit a third network $H_\phi$ to generate the first integral of the system.  Across six benchmark systems (harmonic oscillator, simple and twin pendulums, Lotka--Volterra, pixel pendulum, sphere geodesics) we find each system's relevant symmetries, and the discovered invariants match ground-truth conserved quantities at $|\rho|\geq 0.978$. On the sphere, the pipeline recovers a non-abelian $\mathfrak{so}(3)$, certified by the negative-definite Killing form of its recovered structure constants. This work presents significant advances in data-driven discovery of symmetries and provides a new pathway for neural surrogates of dynamical systems. 
\end{abstract}


\begin{figure}[h]
    \centering
    \includegraphics[width=0.99\linewidth]{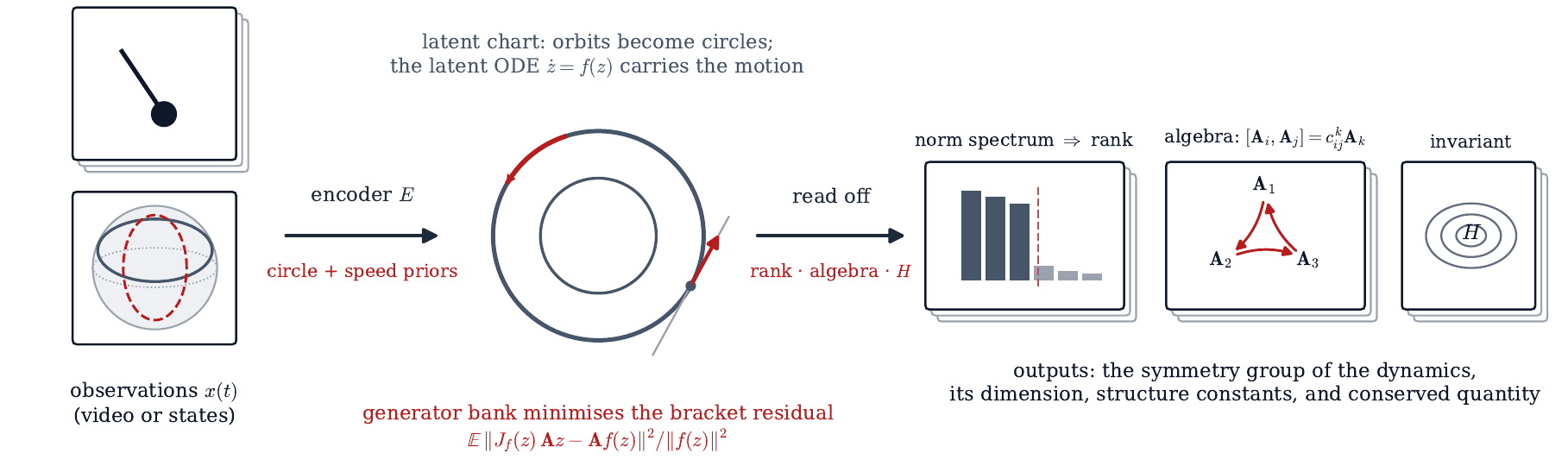}
    \vspace{-3mm}
    \caption{\small An autoencoder maps observations (states or video) to a decorrelated latent chart where a latent ODE $\dot{z} = f(z)$ carries the dynamics and orbits are circular. A linear generator $\mathbf{A}$ is then fitted by minimising the normalised Lie-bracket residual $\mathbb{E}_{z \sim data} \| \mathbf{J}_f(z) \mathbf{A} z - \mathbf{A} f(z) \|^2/\|f(z)\|^2$ on encoded data.}
    \label{fig:schematic}
    \vspace{-6mm}
\end{figure}

\section{Introduction}
\label{sec:intro}

Symmetry is one of the fundamental frameworks of physics. Continuous symmetries are linked with conservation laws via Noether's theorem \citep{Noether1918}, they constrain dynamics, and they reduce the dimensionality of systems. A growing body of work aims to build known symmetries into networks as equivariance constraints \citep{cohen2016group, bronstein_review, villar2021scalars} or to learn dynamics that respect an assumed analytic structure such as a Hamiltonian \citep{greydanus2019hamiltonian, cranmer2020lagrangian, toth2020hamiltonian}. However, the inverse problem is also of great interest: Given only noisy observations of a system, can one determine which symmetries (if any) it has? Existing approaches search for symmetries directly in observation space \citep{liu2021machine, desai2022symmetry} or learn group actions with adversarial objectives \citep{yang2023generative,yang2023latent}, typically on state-space observations with the symmetry group's dimension fixed in advance.\\

We take a different route, motivated by the work of \citet{yang2023latent}. A symmetry that is nonlinear and incomprehensible in observation space is often \emph{linear} in the right coordinate system. Hence, we create a neural pipeline that (1) automatically finds a reduced dimensional representation of a given dynamical system; (2) learns a latent ODE \citep{chen2018neural, rubanova2019latent} from trajectories, mapping observations into a low-dimensional decorrelated coordinate chart; (3) fits a deliberately over-complete bank of linear generators against a differentiable Lie-bracket residual; and (4) reads off the algebra's dimension from the generator-norm spectrum, its structure constants from pairwise brackets, and its conservation laws from jointly-fitted first-integral networks.  

\paragraph{Main contributions:} 
\begin{enumerate}
    \item A fully differentiable symmetry-discovery objective, paired with matrix-norm rank detection, for which the algebra dimension is an output. \vspace{-2mm}
    \item  Discovery and optimisation of a first-integral based on the squared cosine between $\nabla H$ and the flow, invariant to reparameterisation of $H$, with post-hoc calibration to physical units.  \vspace{-2mm}
    \item Validation on five classical system trajectories and a video observation, successfully recovering the true symmetries of each system.  \vspace{-2mm}
\end{enumerate}

\section{Related work}
\label{sec:related}
\paragraph{Noether networks}
 \citep{alet2021noether} meta-learn conserved quantities from raw observations by tailoring a prediction model at test time to respect a learned conservation law, using Noether's theorem as an inductive bias for forecasting. The conserved quantity is discovered with the goal of improving prediction accuracy.

\paragraph{LaLiGAN}
\citep{yang2023latent}, which extends work of \citet{yang2023generative}, learns an autoencoder for a latent space, in which a symmetry becomes a linear group action, representing it by Lie algebra basis matrices. Training is carried out adversarially: A discriminator tests whether group-transformed latents remain indistinguishable from the data distribution. It thereby discovers distribution-level symmetries with the group's dimension fixed in advance, treating dynamics and downstream structure (e.g. latent equation discovery) as separate, subsequent steps.


\paragraph{Hamiltonian and Lagrangian Neural Networks}
\citep{greydanus2019hamiltonian, cranmer2020lagrangian} parameterise the scalar generating function of an assumed variational structure, which obtains the dynamics and exact energy conservation by construction. Autoencoder variants extend both approaches to pixel observations.

\paragraph{Key distinctions}

Hamiltonian and Lagrangian Neural Networks solve the inverse of our problem, with physical structure being imposed, rather than discovered, to learn better dynamics. LaLiGAN is the closest work. It also learns an autoencoder for a latent space where a nonlinear symmetry acts linearly and parameterises the symmetry by Lie algebra basis matrices. LaLiGAN's symmetry condition is enforced distributionally, through a discriminator that judges whether transformed pairs remain indistinguishable from real ones, and is therefore only defined at an adversarial optimum. We impose the infinitesimal Lie bracket condition $[v_A, f] = 0$ directly on the learned vector field and evaluate it pointwise by automatic differentiation, giving a scalar residual at every state. First, the bracket can be thresholded, compared on- versus off-manifold, and reported as a verification statistic. Secondly, in our method the algebra's dimension is an output of the spectral gap rather than a hyperparameter. Finally, we recover first integrals alongside the generators and calibrate them to physical units, a connection between discovered symmetry and conservation laws hinted at as future work by \citet{yang2023latent}.
\vspace{-4mm}

\begin{table}[htbp]
\floatconts
{tab:related}%
{\caption{Capability comparison with the closest families of methods.}}%
{\vspace{-3mm}}
{\resizebox{\textwidth}{!}{%
\begin{tabular}{lcccc}
\toprule
 & Ours & LaLiGAN & Noether Nets & HNN/LNN \\
\midrule
Discovers Lie generators                       & \cmark & \cmark & --     & --     \\
Continuous-time dynamics learned jointly       & \cmark & --     & --     & \cmark \\
Invariants discovered, not assumed             & \cmark & --     & \cmark & --     \\
Num. of symmetries detected, not specified     & \cmark & --     & --     & --     \\
Expressivity guarantee for the action          & --     & \cmark & --     & --     \\
Exact conservation by construction             & --     & --     & --     & \cmark \\
\bottomrule
\end{tabular}}}
\end{table}
\vspace{-7mm}

\section{Method}
\label{sec:method}

\paragraph{Latent dynamics model}  We follow the non-variational latent ODE architecture from \citet{pathmin}, a modification of the implementation from \citet{rubanova2019latent}, which encodes time series with a recurrent neural network. We do not employ the explicit path-minimizing penalty used by \citet{pathmin}, which is designed to perform parameter inference. It would compress the dynamics, while we seek to reshape them. For video/pixel data, the encoder and decoder are small convolutional stacks. Training minimises reconstruction error plus a latent-consistency term; for video, the per-pixel squared error is weighted by each pixel's temporal variance (normalised to a mean of one), which removes the otherwise-dominant trivial optimum of predicting a static background.

\paragraph{Coordinate construction}                                                  The latent chart (our new coordinate system) is recomputed from batch statistics at every training step via:
\vspace{0.4mm}                                                                                                 
\begin{tcolorbox}[colback=black!5!white, colframe=blue!5!white, boxrule=0pt]                                   
\begin{equation}           
    \varphi(z) \;=\; (z - c)\,\mathbf{T}^{\top}, \qquad \mathbf{T} = \mathbf{D}^{\frac{1}{2}} \tilde{\mathbf{C}}^{-\frac{1}{2}} \mathbf{D}^{-\frac{1}{2}}, 
\label{eq:whiten}             
\end{equation}                  
\end{tcolorbox}                                                                                                
\vspace{0.4mm}                                                                                                 
\noindent                                                                                                      
where $c$ re-centres the batch of latents at the origin, $\mathbf{D}$ is the diagonal matrix of per-dimension batch variances, and $\tilde{\mathbf{C}}$ is the batch correlation matrix, whose inverse square root is computed by a Newton--Schulz iteration. $\mathbf{T}$ is therefore a variance-preserving decorrelation. The cross-correlations between latent dimensions are removed, while each dimension's scale is kept, so that symmetries of the dynamics become linearly representable. A small $\varepsilon=1\times10^{-8}$ is added to protect against singularities.

\paragraph{Setting and assumed form} We seek one-parameter groups of transformations $g_s$ of the latent state that commute with the flow $\Phi_t$ of the learned dynamics $\dot z = f(z)$, i.e. $g_s \circ \Phi_t = \Phi_t \circ g_s$, so that every solution maps to another. With $v$ representing the vector field that generates $g_s$, this condition is equivalent to the vanishing of the Lie bracket $[v, f] = 0$. We restrict the search to linear generators $v(z) = \mathbf{A}z$ in a learned coordinate chart, so $g_s = e^{s\mathbf{A}}$. The coordinate chart is trained to carry the nonlinearity of the systems. The decorrelation and the action-angle priors are chosen so that phase rotations, dilations and rigid rotations of (quasi-)periodic systems become linear, and symmetries with no linear representation at the chosen latent width are outside the search space by construction. No Hamiltonian or Lagrangian structure is assumed and $f$ is an unconstrained smooth vector field and $H_\phi$ is a first integral of it.

\paragraph{Rotational dynamics} Apart from decorrelated dimensions, we must explicitly encourage circular trajectories in latent space. This is achieved through a penalty term:
\noindent
\begin{tcolorbox}[colback=black!5!white, colframe=blue!5!white, boxrule=0pt]
\begin{equation}    
  \mathcal{L}_{\mathrm{circle}} \;=\; \frac{\;\mathbb{E}_{i}\!\left[\operatorname{Var}_{t} \left(r_i(t)\right)\right]\;}{\;\mathbb{E}_{i,t}\!\left[r_i(t)^{2}\right] + \varepsilon\;}, \qquad   r_i(t) \;=\; \bigl\lVert \varphi\bigl(z_i(t)\bigr) \bigr\rVert ,         
\label{eq:circle}              
\end{equation}    
\end{tcolorbox}
\noindent
which penalises the temporal variance of the radius $\lVert\varphi(z_i(t))\rVert$ along each trajectory. We normalise by the squared radius to avoid trivially shrinking $r$ to 0. This term vanishes exactly when every orbit is an origin-centred circle of constant radius, so periodic motion acquires an action-like radial coordinate and rotations become linearly representable.

\noindent 
To further encourage action-angle representations, we include a speed penalty: 
\begin{tcolorbox}[colback=black!5!white, colframe=blue!5!white, boxrule=0pt]
\begin{equation}
\mathcal{L}_{\mathrm{speed}} \;=\;\frac{\;\mathbb{E}_{i}\!\left[\operatorname{Var}_{t}\left( s_i(t)\right)\right]\;}{\;\mathbb{E}_{i,t}\!\left[s_i(t)^{2}\right] + \varepsilon\;}, \qquad s_i(t) \;=\; \bigl\lVert \mathbf{T}\, f\bigl(z_i(t)\bigr) \bigr\rVert ,                      
\label{eq:speed}                                                                                                                             
\end{equation}
\end{tcolorbox}
\noindent
applying the same scale-free regularisation to the pushforward field magnitude $\lVert \mathbf{T}f(z_i(t))\rVert$\footnote{where $\mathbf{T}$ is the linear component of the coordinate chart from \autoref{eq:whiten}}, demanding each orbit be traversed at uniform speed on a constant-radius orbit, i.e. at uniform angular velocity. This penalty encourages the true action-angle representation and thus allows for the discovery of linear Lie point symmetries.

\paragraph{Symmetry discovery} We  construct an over-complete bank of $K$ linear infinitesimal generators $\{\mathbf{A}_k\}$, where the generators are fitted by minimising the Lie bracket loss:
\vspace{1mm}
\begin{tcolorbox}[colback=black!5!white, colframe=blue!5!white, boxrule=0pt]
\begin{equation}
\label{eq:bracket}
\mathcal{L}_{b} = \mathbb{E}_{z \sim data}
\frac{1}{K}\sum_{k=1}^{K}\frac{\| \mathbf{J}_f(z) \mathbf{A}_k z - \mathbf{A}_k f(z) \|^2}{\|f(z) \|^2 + \varepsilon },
\end{equation}
\end{tcolorbox}
\vspace{1mm}
\noindent
evaluated by forward-mode differentiation using a Jacobian-vector product (\texttt{jax.jvp}) to avoid a full Jacobian computation. We sample $z$ from encoded real trajectories with a small jitter. Generator redundancy is suppressed by decorrelating the actions $\mathbf{A}_k z$ in the data metric. The number of surviving generators after a multiplicative-gap test on the norm spectrum (see \appendixref{sec:gap-test}) is the detected algebra dimension, and closure is then checked by projecting pairwise brackets back onto the surviving span. 

\paragraph{Independent generators}                                                                            
Training several candidate generators is prone to failures: Without decorrelation pressure they all collapse onto the single easiest symmetry, while a naive decorrelation pressure prevents convergence altogether. We adapt the stop-gradient orthogonalisation of \citet{markov2026datadriven}, where each pair penalty anchors the earlier generator of the pair, so generator~1 settles freely, generator~2 arranges itself to differ from a fixed generator 1, and so on. We measure overlap as the squared correlation of the generator actions ($\mathbf{A}_i z$) on the data manifold:
\begin{tcolorbox}[colback=black!5!white, colframe=blue!5!white, boxrule=0pt]                                   
\begin{equation}                                                                 
\mathcal{L}_\perp \;=\; \sum_{i<j} \frac{\big(\mathbb{E}_z\langle \mathrm{sg}(\mathbf{A}_i z),\, \mathbf{A}_j z\rangle\big)^2}{\mathrm{sg}\!\big[\mathbb{E}_z\|\mathbf{A}_i z\|^2\big]\, \mathrm{sg}\!\big[\mathbb{E}_z\|\mathbf{A}_j z\|^2\big] + \varepsilon}           
\label{eq:orth}
\end{equation}                                                                   
\end{tcolorbox}                                                                                                
\noindent                                                                                               
With latent dimension above the true effective dimension, overlap can be cancelled by matrix components acting on dead dimensions, letting duplicate generators hide. The normalisation and the samples $z$ are also stop-gradiented, so the pressure only steers generator directions and cannot be satisfied by inflating norms or bending the coordinates. 

\paragraph{Joint objective}             
All of the symmetry terms above would be trivially minimised by $\mathbf{A}_k = 0$. We introduce a norm budget to prevent that while still allowing redundant generators to go to 0:
\begin{tcolorbox}[colback=black!5!white, colframe=blue!5!white, boxrule=0pt]
\begin{equation}
\label{eq:budget}
\mathcal{L}_{\mathrm{budget}} \;=\;
\Big(\textstyle\sum_{k=1}^{K} \mathbb{E}_z\|\mathbf{A}_k z\|^{2} \;-\; K\Big)^{2}
\;+\; \gamma\,\frac{1}{K}\textstyle\sum_{k=1}^{K}
\big(\mathbb{E}_z\|\mathbf{A}_k z\|^{2} - 1\big)^{2},
\end{equation}
\end{tcolorbox}
\noindent
where $K$ is the bank size. The first term fixes only the total on-manifold norm; the second (weight $\gamma = 0.05$) is a gentle per-generator attraction toward unit norm, preventing a winner-take-all allocation when several genuine symmetries coexist. Norms are measured through the data as $\mathbb{E}_z\|\mathbf{A}_k z\|^{2}$, so a generator acting only off-manifold counts as collapsed. The full symmetry objective
\begin{tcolorbox}[colback=black!5!white, colframe=blue!5!white, boxrule=0pt]
\begin{equation}
\label{eq:joint-loss}
\mathcal{L}_{\mathrm{Lie}} \;=\; \lambda_1\mathcal{L}_b
+ \lambda_2\mathcal{L}_{\perp}
+ \lambda_3\mathcal{L}_{\mathrm{circle}}
+ \lambda_4\mathcal{L}_{\mathrm{speed}}
+ \lambda_5\mathcal{L}_{\mathrm{budget}}
+ \lambda_6\mathcal{L}_{H\!A}
\end{equation}
\end{tcolorbox}
\noindent
is added to the reconstruction and latent-consistency losses of the dynamics model and optimised jointly in the main training phase; $\mathcal{L}_{H\!A}$ is the Noether-compatibility pressure defined in \autoref{eq:integral} below. The symmetry weights sit well below the fit weights and ramp in over training: at equal weighting, a small gain in bracket residual can buy a large reconstruction degradation, after which
every ``symmetry'' found is a symmetry of model error. Per-system weights $\lambda_i$ are given in \appendixref{sec:hypers}.

\paragraph{Discovering conservation laws}
Independent \textit{first-integral} networks $H_\phi$ are learned alongside the generators. A quantity is conserved exactly when its gradient is perpendicular to the motion, and a symmetry is Noether-compatible (i.e.\ corresponds to a conservation law) when it preserves that quantity. This gives two perpendicularity constraints of the same form, with $\cos^{2}(a,b) = \langle a,b\rangle^{2}/(\|a\|^{2}\|b\|^{2}+\varepsilon)$:
\vspace{1mm}
\begin{tcolorbox}[colback=black!5!white, colframe=blue!5!white, boxrule=0pt]
\begin{equation}
\label{eq:integral}
\mathcal{L}_{H} \;=\; \mathbb{E}_{z}\,\cos^2\!\big(\nabla H_\phi(z),\, f(z)\big),
\qquad
\mathcal{L}_{H\!A} \;=\; \mathbb{E}_{z}\,\cos^2\!\big(\mathrm{sg}[\nabla H_\phi(z)],\, \mathbf{A} z\big).
\end{equation}
\end{tcolorbox}
\vspace{1mm}
\noindent
When a system carries several independent invariants, the networks (multiple heads) additionally receive a decorrelation pressure $\mathcal{L}_{H\perp} = \sum_{a<b}\mathbb{E}_z \cos^2\big(\mathrm{sg}[\nabla H_a(z)], \nabla H_b(z)\big)$, i.e. gradients of distinct heads are pushed away from each other on the data manifold, with the stop-gradient anchoring the earlier head as in $\mathcal{L}_\perp$, so the two factor energies of the twin pendulums are recovered by separate networks rather than collapsing onto one. The left term trains $H$ to be constant along the flow\footnote{We note this term is specifically a first integral, not the Hamiltonian, as it carries no symplectic structure and cannot generate the dynamics.}. Because the cosine is scale-free, $\mathcal{L}_H$ is also minimised by the degenerate solution $\nabla H \equiv 0$, we add a unit-gradient penalty $\mathbb{E}_z\big(\lVert\nabla H(z)\rVert - 1\big)^{2}$ to disallow this solution. The right term steers the generators, the stop-gradient $\mathrm{sg}[\cdot]$ freezes $H$, so the penalty pushes $\mathbf{A}$ toward transformations that slide states along level sets of $H$ rather than across them. Both terms are active during the main phase, so $H_\phi$ co-trains with the dynamics and the bank. Because a fraction of initialisations stall in a $\cos^{2}$ plateau, $H_\phi$ is over-complete (several independent networks); after the main phase the networks are re-fitted against the frozen $f$, $\{\mathbf{A}_k\}$, and coordinate chart by minimising the conservation terms alone, and the best network is kept. We then calibrate the discovered invariant to physical units by a monotone fit on validation data (see \autoref{fig:energy}a).

\section{Results}
\label{sec:results}

\begin{figure}[h]
\includegraphics[width=0.92\linewidth]{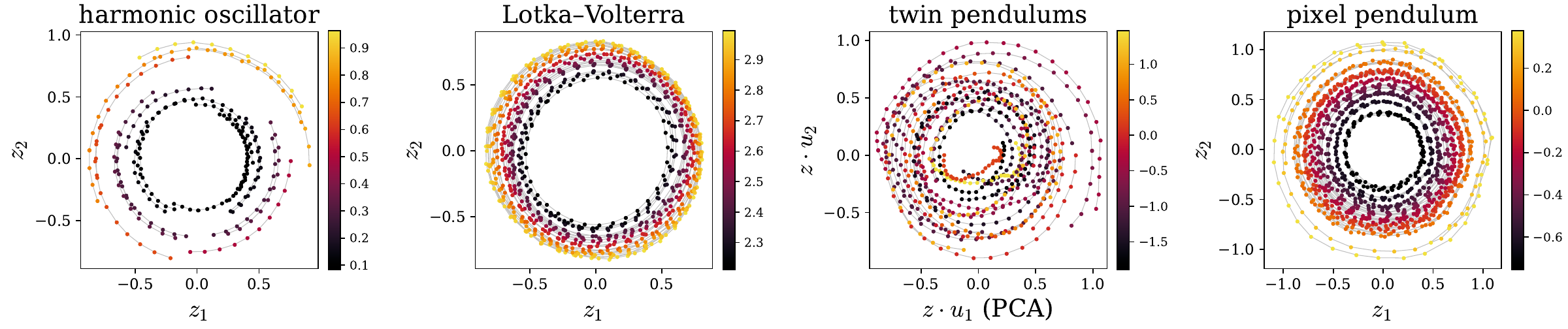}
\vspace{-3mm}
\caption{\small Latent phase portrait of the harmonic oscillator, Lotka-Volterra, twin pendulum, and pixel-pendulum systems. Note that if the reduced latent dimension is $> 2$, a PCA projection is used to convert the $n$-dimensional trajectory onto a 2D map (as in the twin pendulum).}
\label{fig:circles}
\vspace{-5mm}
\end{figure}

\paragraph{Circular dynamics}
\autoref{fig:circles} shows the latent trajectories of four of our test systems. The colourbar can be considered to represent the unit-less energy of each trajectory. We see that the autoencoder is able to create almost entirely circular trajectories even for highly nonlinear systems (twin pendulums) or pure pixel space observations (pixel pendulum).

\paragraph{Symmetry discovery}
\input{table1}

Across all six systems, our \textsc{LatentLieODE} method recovers the expected symmetry (see \autoref{tab:results}). The discovered invariant for each system also matches the ground-truth conserved quantity (correlation coefficient $|\rho|$ in \autoref{tab:results}) almost perfectly. 
Wherever dimension discovery is enabled, we see successful reduction of the latent dimensions to the lowest true dimensionality of each respective system\footnote{We explicitly set the latent dimensionality for both the twin pendulums and sphere geodesics ($\mathfrak{so}(3)$ requires the extra slack despite the system requiring only 4 dimensions).}, and we find the true number of symmetries for each system (rank) despite never specifying the desired amount a priori. 

\begin{figure}
\includegraphics[width=0.9\linewidth]{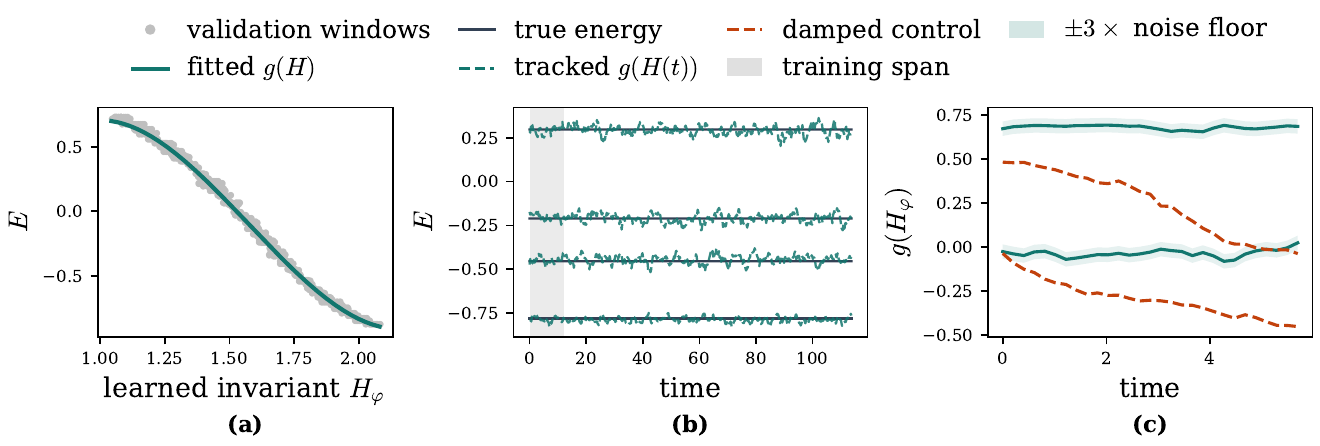}
\vspace{-3mm}
\caption{\small The discovered invariant, calibrated to energy on validation windows \textbf{(a)}, tracks the true energy of held-out trajectories over $10\times$ the training span \textbf{(b)}, and yields a falsifiable conservation verdict \textbf{(c)}: fresh pendulum data stays within the $\pm3\times$ noise-floor band ($1.6\times$ the floor), while a damped control from the same initial conditions is correctly rejected at $13\times$ the noise floor.}
\label{fig:energy}
\vspace{-6mm}
\end{figure}

\paragraph{Transferring the discovered invariant}
The discovered $H$, its coordinate chart, and its calibration together define a standalone function on raw observations, so conservation can be tested on data the model has never seen. \autoref{fig:energy}a shows the monotone calibration map fitted on validation windows. Since the $\cos^2$ objective is invariant under reparameterisation of $H$, this monotone fit converts the invariant to physical units. On unseen pendulum trajectories, the calibrated invariant tracks the true energy ($|\rho| = 0.998$ from \autoref{tab:results}) and remains conserved for over ten times the training data period (\autoref{fig:energy}b). Importantly, we can test this learned invariant on systems that do not conserve energy, e.g. a damped pendulum (\autoref{fig:energy}c). In this case, we see a consistent drop in $H$ along the observed trajectories, indicating a clear violation of energy conservation for the system.

\begin{figure}[t]
\includegraphics[width=0.99\linewidth]{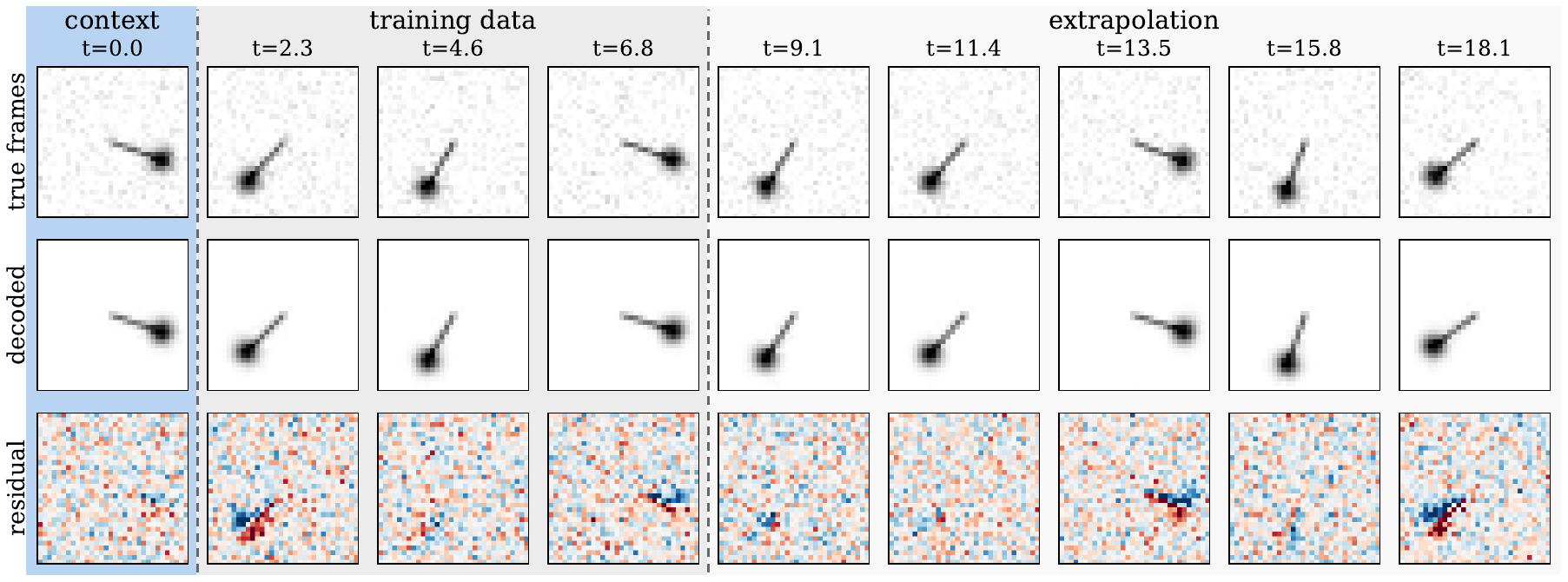}
\vspace{-3mm}
\caption{\small Pixel-pendulum reconstruction. Ground-truth frames \textbf{(top)} vs. frames decoded from the \textsc{LatentLieODE} rollout \textbf{(middle)} and residuals (truth minus decoded, \textbf{bottom}), continued past the end of the training data. All temporal structure lives in the 2D latent ODE with the decoder rendering each latent state independently.}
\vspace{-6mm}
\label{fig:pendulums}
\end{figure}

\paragraph{Pixels to physics}
\label{sec:pixels}
For the pixel pendulum, the model receives only $32\times32$ pixel frames with no state supervision. The initial dimensionality reduction prunes the latent space from 8 dimensions to 2, where we then run the main symmetry discovery step. From \autoref{tab:results} we can see \textsc{LatentLieODE} finds one symmetry of this system, with the discovered invariant tracking the true pendulum energy at $|\rho| = 0.995$ (calibrated $R^2 = 0.996$; energy error $1.7\%$, and $1.6\%$ drift on fresh trajectories). \autoref{fig:pendulums} shows the resulting model rolled out and decoded into observation space from a single context window to twice the training span. We see strong agreement between the true and reconstructed dynamics far into the extrapolated regime.

\begin{figure}[t]
    \centering
    \includegraphics[width=0.99\linewidth]{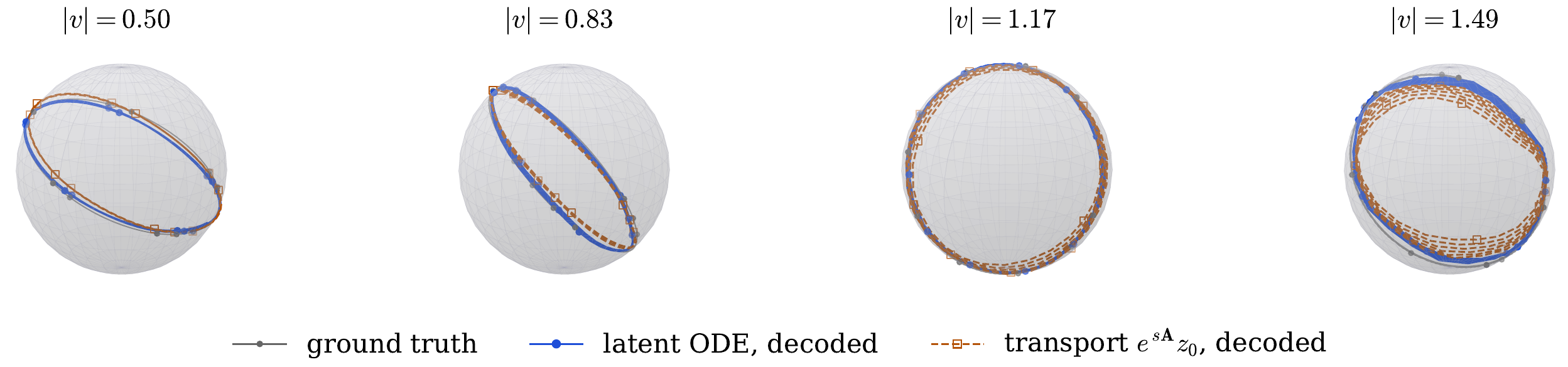}
    \vspace{-3mm}
    \caption{\small Ground truth (gray), decoded latent-ODE rollout (blue), and
    pure group transport $e^{s\mathbf{A}}z_0$ (orange, dashed; no integration,
    $s = t$) for four held-out orbits. The transport generator is the per-orbit projection of the flow onto the discovered $\mathfrak{so}(3)$ span.}
    \label{fig:sphere}
    \vspace{-6mm}
\end{figure}

\paragraph{Non-abelian symmetry groups}
For all preceding systems the discovered algebras are abelian, so closure is trivial. As a non-abelian test, we simulate free-particle geodesics on $S^2$, specifically, state observations $(x,v) \in \mathbb{R}^6$, with particle speeds drawn from $|v| \sim \rm{Uniform}(0.5, 1.5)$ in units of a sphere radius 1. The true algebra of this system is $\mathfrak{so}(3)$. We recover exactly three symmetries from three surviving generators. This system is visualised in \autoref{fig:sphere}, where we show trajectories of increasing particle speeds from left to right. 

Projecting the pairwise commutators of the three surviving generators back onto their span recovers the bracket table
\begin{equation*}
  [\mathbf{A}_1,\mathbf{A}_2] = 1.82\,\mathbf{A}_3, \qquad
  [\mathbf{A}_2,\mathbf{A}_3] = 1.72\,\mathbf{A}_1, \qquad
  [\mathbf{A}_3,\mathbf{A}_1] = 1.65\,\mathbf{A}_2.
\end{equation*}
The Killing form assembled from these structure constants, $K_{ij} = \sum_{m,n} c^m_{in}c^n_{jm}$, has eigenvalues
$(-6.25, -5.98, -5.66)$, hence is negative definite. Definiteness is invariant under change of basis, and among three-dimensional real Lie algebras a negative-definite Killing form identifies $\mathfrak{so}(3)$ uniquely. As a consistency check, an orthonormal $\mathfrak{so}(3)$ basis scaled by $\lambda$ gives $K = -2\lambda^2 I$, and the observed eigenvalues imply $\lambda \approx 1.7$, closely matching the recovered structure-constant magnitudes of 1.82, 1.72 and 1.65 (see \appendixref{apd:killing-scale} for more details). From the run's stored per-pair projection residuals and commutator norms, the off-span fractions $|P_\perp[\mathbf{A}_i,\mathbf{A}_j]|_F /|[\mathbf{A}_i,\mathbf{A}_j]|_F$ are:
\begin{equation*}
(\mathbf{A}_1,\mathbf{A}_2): 1.8\%, \qquad (\mathbf{A}_1,\mathbf{A}_3): 1.6\%, \qquad (\mathbf{A}_2,\mathbf{A}_3): 2.2\%
\end{equation*}
i.e., $\geq97.8\%$ of every commutator lies inside the surviving span.

\paragraph{Integration-free dynamics}
\figureref{fig:sphere} shows trajectories from the integration of $f$ as well as the symmetry-evolved dynamics for the particles on a sphere. In both cases, initial states $z_0$ are encoded from held-out trajectory windows, but instead of ordinarily integrating the ODE (blue curves), the dashed orange curves apply the discovered generator via the exponential map $e^{s\mathbf{A}}z_0$. These dynamics are purely generated by increasing the values of $s$. The close agreement between all three curves is further evidence that we have indeed discovered valid symmetries for this physical system.

\section{Discussion}
\label{sec:discussion}
We showed that a latent ODE equipped with a decorrelated coordinate chart, action-angle-encouraging penalties, and a differentiable bracket residual discovers linear Lie symmetries, their algebra dimension, structure constants, and calibrated first integrals from state timeseries and from video frames. We find strong agreement between the rollout of the latent trajectories whether using direct numerical integration of the ODE function $f$ or via sweeping the action parameter $s$ of the exponential map to create the temporal dynamics. We have thus demonstrated significant advances in data-driven discovery of symmetries and provide a new pathway for neural surrogates of dynamical systems.

\paragraph{Limitations} 
The method returns linear generators in a learned chart, so a symmetry is found only if some chart at the chosen latent width linearises it. The circle and speed priors presume (quasi-)periodic orbits, so chaotic regimes are out of scope, and results are single training runs checked by scripted acceptance gates rather than seed ensembles.

\bibliography{paper}

\clearpage
\appendix

\section{Cost and scaling}
\begin{figure}[h]
      \centering
      \includegraphics[width=0.8\linewidth]{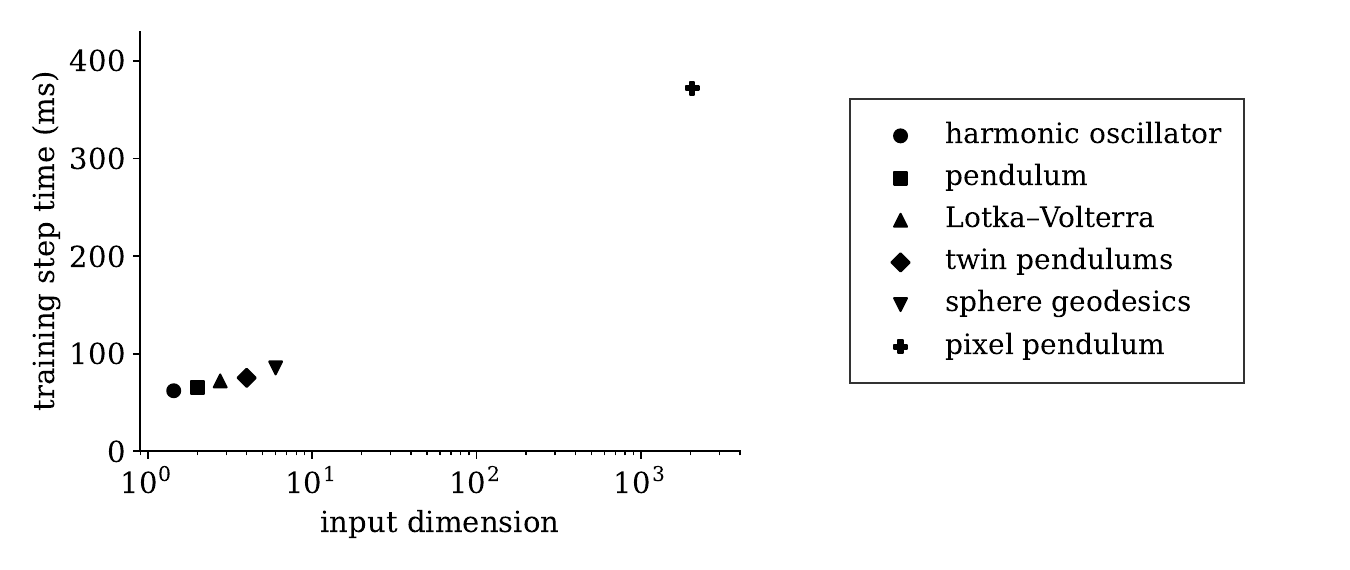}
      \caption{\small Training step time against encoder input dimension, single CPU.}
      \label{fig:timing}
  \end{figure}

  The symmetry machinery scales with the latent width $d$, not the observation dimension $D$. The bracket residual costs one Jacobian-vector product per sample
  and generator, an extra forward pass of $f$ with no Jacobian materialised, and the chart is a $d \times d$ batch statistic. \figureref{fig:timing} shows this and we see step time is mostly flat across $D = 2 \to 6$, with the pixel-pendulum at $D=2048$ costing roughly 4 times more than the other systems. 

\section{Sphere geodesic Killing form}
\label{apd:killing-scale}
This section derives the consistency check quoted in the main text, the eigenvalues of the Killing form and the magnitudes of the recovered structure constants. 

\noindent
Throughout, indices $i, j, k, m, n$ run over the three surviving generators $\mathbf{A}_1, \mathbf{A}_2, \mathbf{A}_3$. The structure constants $c^k_{ij}$ are the projection coefficients of the pairwise commutators onto
the surviving span,
\begin{equation}
    [\mathbf{A}_i, \mathbf{A}_j] \;=\; \sum_{k} c^k_{ij}\,\mathbf{A}_k ,
\label{eq:structure-constants}
\end{equation}
and the Killing form is the symmetric $3\times3$ matrix built from them:
\begin{equation}
    K_{ij} \;=\; \sum_{m,n} c^m_{in}\, c^n_{jm}
    \;=\; \mathrm{tr}\!\left(\mathrm{ad}_i\,\mathrm{ad}_j\right),
    \qquad
    (\mathrm{ad}_i)^k_{\;j} = c^k_{ij} .
\label{eq:killing-def}
\end{equation}
$K$ depends only on the structure constants.

\paragraph{Baseline: the orthonormal $\mathfrak{so}(3)$ basis.}
For an orthonormal basis $X_1, X_2, X_3$ of $\mathfrak{so}(3)$ the
structure constants are given by,
\begin{equation}
    [X_i, X_j] \;=\; \sum_k \varepsilon_{ijk}\, X_k ,
    \qquad
    \varepsilon_{123} = \varepsilon_{231} = \varepsilon_{312} = +1,
    \quad
    \varepsilon_{213} = \varepsilon_{132} = \varepsilon_{321} = -1,
\label{eq:so3-basis}
\end{equation}
(all other components zero). Substituting
$c^k_{ij} = \varepsilon_{ijk}$ into \equationref{eq:killing-def} and using
the contraction identity
$\sum_{m,n}\varepsilon_{inm}\,\varepsilon_{jmn} = -2\,\delta_{ij}$
(where $\delta_{ij}$ is the Kronecker delta) gives
\begin{equation}
    K \;=\; -2\,I .
\label{eq:killing-orthonormal}
\end{equation}

\paragraph{A uniformly scaled basis.}
The basis returned by rank detection is arbitrary in scale. Suppose that every generator carries the \emph{same} unknown scale $\lambda > 0$,
\begin{equation}
    \mathbf{A}_i \;=\; \lambda\, X_i .
\label{eq:uniform-scale}
\end{equation}
The commutator is bilinear, so
\begin{equation}
    [\mathbf{A}_i, \mathbf{A}_j]
    \;=\; \lambda^2\, [X_i, X_j]
    \;=\; \lambda^2 \sum_k \varepsilon_{ijk}\, X_k
    \;=\; \lambda \sum_k \varepsilon_{ijk}\, \mathbf{A}_k ,
\label{eq:scaled-bracket}
\end{equation}
i.e.\ the structure constants of the scaled basis are
$c^k_{ij} = \lambda\,\varepsilon_{ijk}$. Since
\equationref{eq:killing-def} is quadratic in the structure constants,
\begin{equation}
    K \;=\; -2\lambda^2\, I .
\label{eq:killing-scaled}
\end{equation}
Each eigenvalue $\kappa$ of the \emph{measured} Killing form therefore
yields an estimate of the basis scale:
\begin{equation}
    \lambda \;=\; \sqrt{-\kappa / 2\,} .
\label{eq:lambda-from-kappa}
\end{equation}
The three measured eigenvalues, computed from the recovered structure
constants of the sphere-geodesic run, are
$\kappa = (-6.25,\; -5.98,\; -5.66)$, giving
\begin{equation}
    \lambda \;=\;
    \left(\sqrt{6.25/2},\; \sqrt{5.98/2},\; \sqrt{5.66/2}\right)
    \;=\; (1.77,\; 1.73,\; 1.68) .
\label{eq:lambda-values}
\end{equation}

\section{Generator norm ranking}
\label{sec:gap-test}
\begin{figure}[h]
    \centering
    \includegraphics[width=0.7\linewidth]{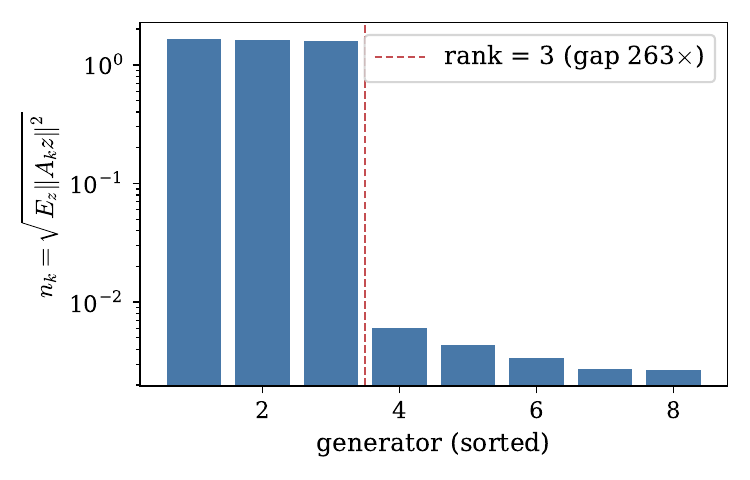}
    \caption{\small Generator norm spectrum for the sphere-geodesic system of $K=8$ deliberately over-complete generators, three survive and five collapse, leaving a $263\times$ gap. The detected algebra dimension matches the expected rank of 3 for the $\mathfrak{so}(3)$ group.}
    \label{fig:sphere_generators}
\end{figure}

\begin{figure}[h]
    \centering
    \includegraphics[width=0.98\linewidth]{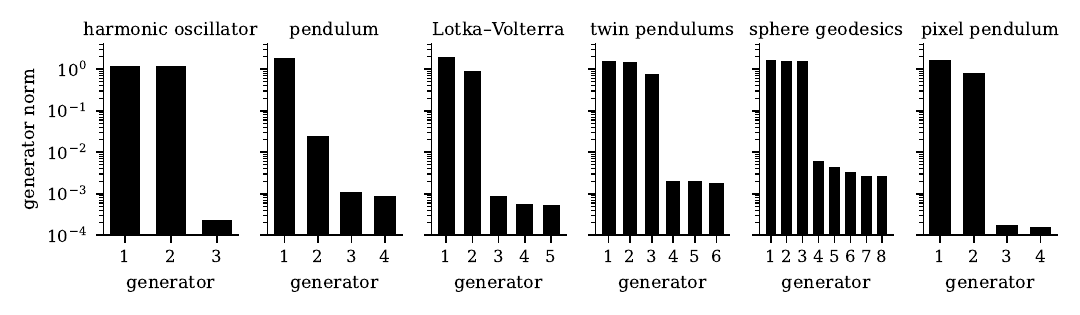}
    \caption{\small Generator norm spectra for the six systems of \autoref{tab:results}, in the same form as \figureref{fig:sphere_generators}}
    \label{fig:spectra}
\end{figure}

To determine the number of symmetries of a certain system, post training we rank each infinitesimal generator $\mathbf{A_k}$ by its norm. We show an example of this in \figureref{fig:sphere_generators} for the sphere geodesic test, where we can see we have 3 surviving generators, corresponding to 3 independent symmetries discovered. We show the generator plots for all the test cases in \figureref{fig:spectra}.

\section{Experimental details}
\label{sec:hypers}

\paragraph{Latent ODE architecture}
The encoder is a reverse-time GRU over $[t, x]$ context windows, followed by a small GELU MLP head producing the latent initial condition $z_0$. The vector field is $f(z) = \sigma_s \cdot \mathrm{MLP}(z)$ with softplus activations and $\tanh$ output using smooth activations only, so the Jacobian in the bracket residual is differentiable everywhere. We note $\sigma_s$ is a learned scale parameter. The decoder is an MLP applied pointwise to each latent state. For pixel observations the encoder and decoder are small convolutional stacks (channels $16/32/64$) over $32\times32$ frames. Latent rollouts use \texttt{diffrax} (Tsit5, adaptive PID controller, $\mathrm{rtol}=\mathrm{atol}=10^{-5}$). We train with AdamW (weight decay $10^{-4}$, gradient clip $1.0$) with a cosine warmup schedule, with a separate Adam optimiser for the generator bank at learning rate $\eta_{\mathrm{gen}}$. Training runs a dimension-discovery (reduction) warmup, then joint co-training in which the symmetry weights follow a quadratic ramp reaching full strength at $75\%$ of the run. Reconstruction is trained over an increasing horizon of the training data, and the Lie bracket samples are drawn from encoded trajectories plus Gaussian jitter at $5\%$ of the latent spread. The Latent ODE architecture is identical across all datasets.

\paragraph{Harmonic oscillator}
Closed-form trajectories of $\dot q = p$, $\dot p = -q$ with radii drawn uniformly from $[0.4, 1.4]$ and uniform phases; 1024 training and 128 validation trajectories of 50 samples over $t \in [0, 8]$, observation noise $\sigma = 0.01$. Dimension discovery prunes the latent space $6 \to 2$.

\paragraph{Simple pendulum}
$\ddot\theta = -(g/L)\sin\theta$ with $g/L = 1$, integrated numerically; amplitudes drawn uniformly from $[0.5, 2.4]$ radians with 60 samples over $t \in [0, 12]$, observation noise $\sigma = 0.05$. Dimension discovery prunes $6 \to 2$.

\paragraph{Twin pendulums}
Two uncoupled pendulums with $g/L = 1$ and $2$ observed jointly as a 4-dimensional state, amplitudes drawn as above, $\sigma = 0.02$. The latent width is fixed at 4.

\paragraph{Lotka--Volterra}
$\dot x = x - xy$, $\dot y = xy - y$ (all rates unity); initial conditions sampled inside the box $[0.25, 3.2]^2$ filtered to invariant values in $[2.15, 3.0]$; 70 samples over $t \in [0, 14]$, $\sigma = 0.01$. Dimension discovery prunes $20 \to 2$.

\paragraph{Pixel pendulum}
The pendulum simulator rendered as $32\times32$ frames (rod and bob, amplitudes uniform in $[0.5, 2.0]$, pixel noise $\sigma = 0.05$), with a temporal-difference channel appended so the state is per-frame observable.
Training is staged. \emph{Stage 1} (perception only, all symmetry weights zero): convolutional codec with an autoencoder warm-start of 5000 steps, variance-weighted reconstruction, latent width 8, 45{,}000 steps at
learning rate $3\times10^{-3}$. \emph{Stage 2}: the full symmetry pipeline on the frozen stage-1 encoder's decorrelated latent trajectories (observation dimension 8); dimension discovery prunes $8 \to 2$, and a metric-anneal curriculum (held for the first $10\%$ of training, annealed to the plain metric by $40\%$) locks the single-cover chart; 80{,}000 steps.

\paragraph{Sphere geodesic}
Closed-form circle geodesics of a free particle on the unit sphere, observed as the state $(x, v) \in \mathbb{R}^6$; initial positions uniform on $S^2$, tangent directions uniform, speeds $|v| \sim \mathrm{Uniform}(0.5, 1.5)$; 3000 training trajectories of 125 samples over $t \in [0, 25]$, $\sigma = 0.01$. The latent width is fixed at 6 with the sparsity weight zero: the linear $\mathfrak{so}(3)$ realisation lives in $\mathbb{R}^6$ while the data manifold is only 4-dimensional, so the embedding slack is required. After co-training, a field-only polish phase freezes everything except the vector field and fine-tunes full-horizon reconstruction for 25{,}000 steps at learning rate $3\times10^{-4}$, sharpening the frequency profile without touching the chart or the generators.

\begin{table}[t]
\floatconts
{tab:hypers}
{\caption{Per-system hyperparameters. Loss weights are the $\lambda_i$ of the joint objective; reconstruction and latent-consistency weights are $1$ throughout.  Pixel stage 1 trains perception only (all symmetry weights zero) and is omitted; its settings appear in the text.}}
{\resizebox{\textwidth}{!}{
\begin{tabular}{lcccccc}
\toprule
 & Harmonic & Pendulum & Lotka--Volterra & Twin pend. & Pixel (stage 2) & Sphere \\
\midrule
trajectories (train/val)      & 1024/128 & 1024/128 & 1024/128 & 1024/128 & 1024/32 & 3000/128 \\
samples $\times$ span         & $50 \times 8$ & $60 \times 12$ & $70 \times 14$ & $60 \times 12$ & --- & $125 \times 25$ \\
noise $\sigma$                & 0.01 & 0.05 & 0.01 & 0.02 & --- & 0.01 \\
obs.\ dim                     & 2 & 2 & 2 & 4 & 8 & 6 \\
latent dim (found)            & $6 \to 2$ & $6 \to 2$ & $20 \to 2$ & 4 & $8 \to 2$ & 6 \\
generators $K$                & 3 & 4 & 5 & 6 & 4 & 8 \\
$H$ heads                     & 4 & 4 & 4 & 4 & 4 & 4 \\
context length                & 40 & 32 & 35 & 32 & 12 & 40 \\
batch size                    & 16 & 16 & 16 & 16 & 16 & 32 \\
lr / lr$_{\mathrm{gen}}$      & $10^{-2}$/$3{\cdot}10^{-2}$ & $10^{-2}$/$3{\cdot}10^{-2}$ & $10^{-2}$/$3{\cdot}10^{-2}$ & $10^{-2}$/$3{\cdot}10^{-2}$ & $5{\cdot}10^{-3}$/$10^{-2}$ & $5{\cdot}10^{-3}$/$10^{-2}$ \\
co-train / polish steps       & 25k/5k & 25k/5k & 25k/5k & 40k/5k & 80k/6k & 100k/10k \\
\midrule
$\lambda_{\mathrm{bracket}}$  & 0.1 & 0.2 & 0.1 & 0.1 & 0.1 & 0.3 \\
$\lambda_{\perp}$             & 0.1 & 0.2 & 0.1 & 0.1 & 0.1 & 0.1 \\
$\lambda_{\mathrm{circle}}$   & 1.0 & 2.0 & 1.0 & 1.0 & 1.0 & 1.0 \\
$\lambda_{\mathrm{speed}}$    & 1.0 & 2.0 & 1.0 & 1.0 & 1.0 & 1.0 \\
$\lambda_{\mathrm{budget}}$   & 0.05 & 0.1 & 0.1 & 0.1 & 0.1 & 0.1 \\
$\lambda_{H\!A}$              & 0.1 & 0.2 & 0.1 & 0.1 & 0.1 & 0.1 \\
$\lambda_{H}$ (+ grad.\ norm) & 1.0 & 1.0 & 0.2 & 0.2 & 0.2 & 0.2 \\
$\lambda_{\mathrm{sparsity}}$ & 0.05 & 0.15 & 0.05 & -- & 0.05 & -- \\
$\lambda_{H\perp}$            & -- & -- & -- & 0.5 & -- & -- \\
\bottomrule
\end{tabular}}}
\end{table}

\end{document}

%% file: table1.tex
\begin{table}[t]
\centering
\caption{\small Symmetry discovery across the test suite. \textbf{Rank} counts
the survivors of the generator-norm spectrum (i.e.\ the number of discovered
symmetries). \textbf{Expected} is the true number of symmetries. $|\rho|$ is the absolute Spearman rank correlation between the discovered invariant network $H_\phi$ and the ground-truth invariant, both evaluated on held-out validation windows. \textbf{Calib.} $R^2$ is the variance of the ground-truth invariant explained by $g(H_\phi)$, where $g$ is a monotone map fitted on validation windows (\autoref{fig:energy}a) }
\label{tab:results}
\resizebox{\linewidth}{!}{%
\begin{tabular}{lrccccc}
\toprule
System & obs.\ dim & rank & expected & discovered symmetries & $|\rho|$ & calib.\ $R^2$ \\
\midrule
Harmonic oscillator & 2 & 2 & 2 & rotation $+$ dilation & 0.999 & 0.999 \\
Pendulum ($\sigma{=}5\%$) & 2 & 1 & 1 & phase rotation & 0.998 & 0.997 \\
Lotka--Volterra & 2 & 1 & 1 & phase rotation & 0.991 & 0.983 \\
Twin pendulums & 4 & 2 & 2 & two phase rotations & 0.978 & 0.963 \\
Sphere geodesics & 6 & 3 & 3 & $\mathfrak{so}(3)$ rotations & 0.997 & 0.996 \\
Pixel pendulum & $2{\times}1024^\dagger$ & 1 & 1 & phase rotation & 0.995 & 0.996 \\
\bottomrule
\end{tabular}%
}
\par\smallskip
\begin{minipage}{0.95\linewidth}
{\footnotesize\linespread{0.85}\selectfont
$^\dagger32\times32$ rendered frames (1024 pixels); a temporal-difference channel computed from the video itself is appended, so the encoder input is $2\times1024 = 2048$. The channel affects encoding only. Decoding and rollout render each latent state independently. \par}
\end{minipage}
\vspace{-3mm}
\end{table}